\documentclass{article}
\usepackage{spconf,amsmath,graphicx}

\usepackage[skip=1pt,font+=smaller]{subcaption}
\usepackage{booktabs} 
\usepackage{multirow}
\usepackage{makecell}
\usepackage{xcolor}         
\usepackage{color}
\usepackage{amssymb}
\usepackage[]{nicefrac}

\title{Frequency and Scale Perspectives of Feature Extraction}
\name{Liangqi Zhang$^1$, Yihao Luo$^{2,1}$, Xiang Cao$^{3,1}$, Haibo Shen$^1$, Tianjiang Wang$^{1 ^\ast}$ \thanks{$^\ast$ Corresponding author}}
\address{$^1$Huazhong University of Science and Technology \\
$^2$Yichang Testing Technique Research Institute\\
$^3$Changsha University}
\begin{document}
%
\maketitle
\begin{abstract}
	Convolutional neural networks (CNNs) have achieved superior performance 
	but still lack clarity about the nature and properties of feature extraction.
	In this paper, by analyzing the sensitivity of neural networks to frequencies and scales,
	we find that neural networks not only have low- and medium-frequency biases but also prefer different frequency bands
	for different classes, and the scale of objects influences the preferred frequency bands.
	These observations lead to the hypothesis that neural networks must learn the ability to extract features at various scales and frequencies.
	To corroborate this hypothesis, we propose a network architecture based on Gaussian derivatives, 
	which extracts features by constructing scale space and employing partial derivatives as local feature extraction operators to separate high-frequency information.
	This manually designed method of extracting features from different scales allows our GSSDNets to achieve comparable accuracy with vanilla networks on various datasets.
\end{abstract}
\section{Introduction}
\label{sec:intro}
Although CNNs have shown considerable promises in various applications since AlexNet\cite{Krizhevsky2012}, they still lack the necessary interpretability.
In recent years, many specialized interpretation methods have been proposed to explain CNNs.
It seems to be a general consensus that CNNs can extract features of different levels through feature visualization methods\cite{Erhan2009,Yosinski2015} and attribution methods\cite{Zeiler2014, Simonyan2014, Springenberg2015,Smilkov2017}.
Units at lower layers detect concrete patterns such as textures while units at high layers detect more semantically meaningful concepts such as faces\cite{Yosinski2015}.

However, the features (or concepts) extracted by CNNs do not always look understandable to humans and are easily disturbed by small perturbations,
the perturbed ones are called adversarial examples\cite{Szegedy2014, Goodfellow2015}.
Previous works investigated this from the frequency perspective, and \cite{Rahaman2019,Xu2019} found that CNNs favor low frequencies and learn them first.
Moreover, \cite{Wang2020} thought that CNNs also could exploit the high-frequency image components that are not perceivable to humans,
and these high-frequency features would improve the generalization of the network.
It seems like each frequency has a role to play.
To further study the behavior of neural networks and the properties of features,
we analyze the behavior of feature extraction from both frequency and scale perspectives.
At a specific resolution, the frequencies and scales of the objects in the image are often correlated.
The coarser the scale of the same object, the lower the frequency.
However, objects of different classes have different scales, and the scale of the same object also varies greatly.
Therefore, we conduct frequency and scale analyses for different classes and samples,
and we find that different classes not only have different frequency biases but also their frequency biases vary with scale.

Since neural networks do not have the scale and frequency priors and real-world objects are composed of different structures at different scales,
it is necessary to extract the features under all scales and frequencies.
Inspired by the scale-space theory\cite{Koenderink1984,Lindeberg1994},
we employ the Gaussian blurring operator to construct the scale-space representation and use partial derivatives to extract the high-frequency information at each scale,
and the derivatives are also regarded as local feature extraction operators.
Moreover, modern efficient CNNs are usually built on depthwise separable convolutions (DSCs)\cite{SIfre2014},
which factorize a standard convolution into a lightweight depthwise convolution for \textit{spatial filtering} and a 1$\times$1 convolution called a heavier pointwise convolution for feature generation.
This makes it very easy to apply Gaussian derivatives to the depthwise convolutions for manipulating all spatial filtering.
Based on these, we propose the GSSDNet to corroborate the hypothesis that \textit{extracting features at different scales and frequencies is a fundamental ability of CNNs}.
Our main contributions are summarized as follows:
\begin{itemize}
	\item On the entire ImageNet test set, the features for image classification are mainly extracted from the low- and middle-frequency bands,
	      and the lack of high-frequency information only has little impact on the network.
	      The high-accuracy networks are superior to the low-accuracy networks mainly in the low- and middle-frequency bands.
	\item Different classes prefer different frequency bands.
	      The lower and narrower the dependent frequency band, the easier the class is to achieve high accuracy.
	\item The frequency biases of samples are affected by the scale of the classification object.
	      The larger the scale, the lower the frequency bias.
	\item We propose the hypothesis that neural networks must have the ability to extract features at different scales and frequencies.
	      As a consequence, we integrate scale-space derivatives into the GSSDNets to verify this hypothesis and
	      GSSDNets achieve similar accuracy with vanilla networks on various datasets.
\end{itemize}

\section{Frequency and Scale Sensitivity}

\begin{figure}[t]
	\centering
	\small
	\includegraphics[width=\linewidth]{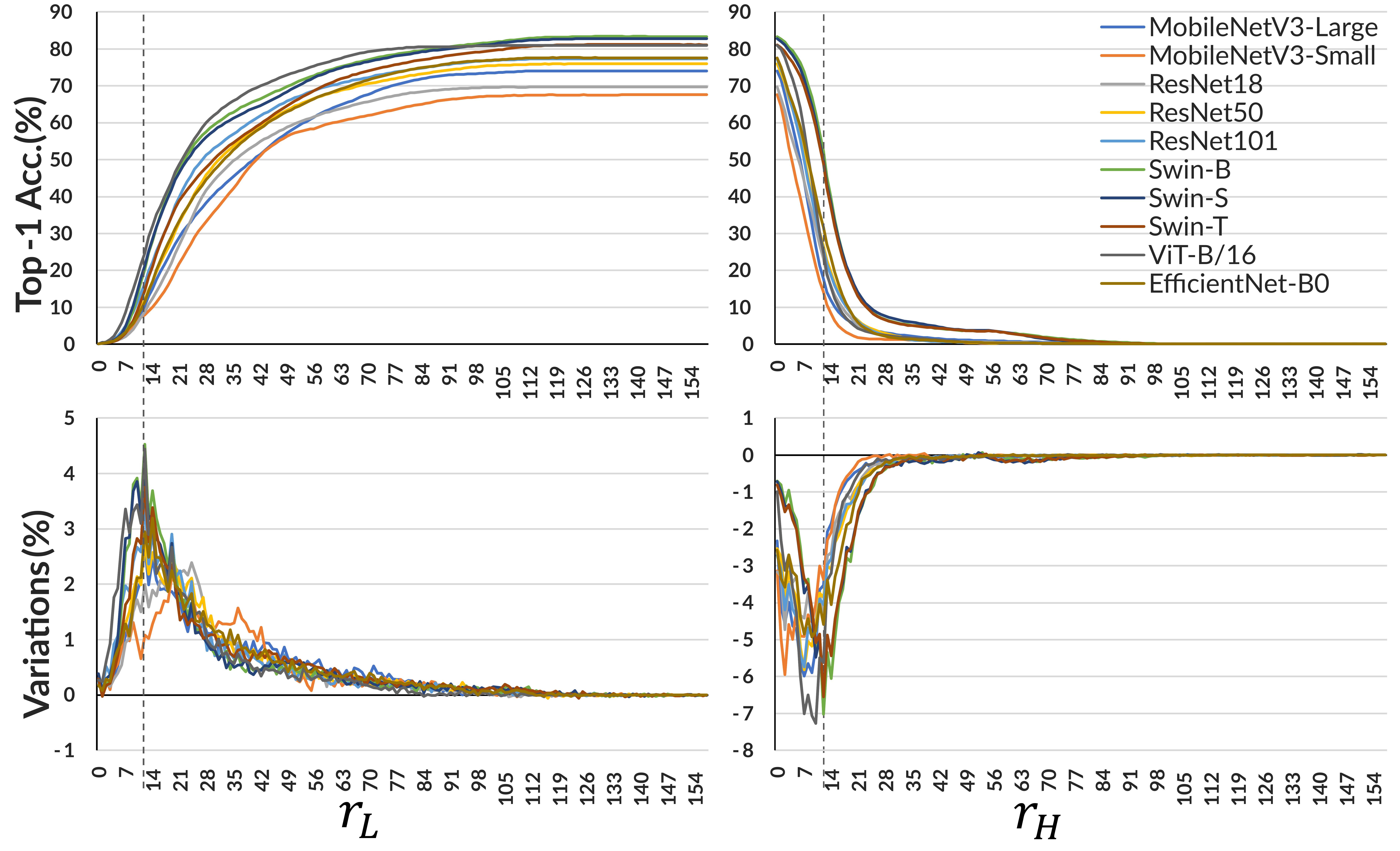}
	\setlength{\abovecaptionskip}{-10pt}
	\caption{Frequency sensitivity analysis.
		Low-frequency bands are more important than high-frequency bands.
		The accuracy of the network is pulled apart mainly in the low- and medium-frequency bands.
	}
	\label{fig:spectral_radius}
\end{figure}
For a more comprehensive analysis of the effects of different frequencies and scales,
this section analyzes the sensitivity of neural networks at the \textit{overall}, \textit{class}, and \textit{sample} levels.
Therefore, we transform images to the frequency domain and then use ideal
low-pass filters (LPF) and high-pass filters (HPF) to remove high-frequency and low-frequency information, respectively.
Then all the experiments are conducted on ImageNet\cite{Deng2010}.
The tested models include both CNNs\cite{He2016, Howard2017, Howard2019, Tan2019a} and transformer-based networks\cite{Dosovitskiy2020, Liu2021},
and all the tested images are at the resolution of $224\times224$.

\subsection{Overall Level}
We tested the effects of missing high- or low-frequency information on the pre-trained models
including ResNets\cite{He2016}, MobileNetV3\cite{Howard2019}, EfficientNets\cite{Tan2019a}, ViT\cite{Dosovitskiy2020} and Swin\cite{Liu2021}.
As shown in Fig. \ref{fig:spectral_radius},
with the increase of LPF radius $r_{L}$, more and more high-frequency information is present,
and the accuracy rises slowly as ultra-low-frequency($r_{L} < 7$) and high-frequency($r_{L} \geqslant 56$) information grows,
but quickly as low-frequency($7 \leqslant r_{L} < 28$) and medium-frequency($28 \leqslant r_{L} < 56$) information does.
With the increase of HPF radius $r_{H}$, the low-frequency information in the image becomes less and less, and the accuracy decreases rapidly before $r_{H} = 28$.
These indicate that the features extracted by the neural network are mainly derived from the low- and medium-frequency information.
In addition, by observing the variations of accuracy,
it can be found that networks with higher accuracy achieve better performance in the low- and middle-frequency bands,
and it is difficult for networks to make correct predictions only based on high-frequency information.

Since directly removing the frequency information from the images in the test set would lead to inconsistent distribution between test and training data,
we preprocess the data with ideal low-pass filters and retrain the model from scratch on ImageNet training data.
As shown in Table \ref{tab:lpf_hpf},
the retrained model achieves much higher accuracy than the pre-trained model at the same $r_{L}$.
This indicates that although the neural network will try to use the information of all frequency bands to achieve the highest accuracy,
it does not make full use of the low- and medium-frequency information (or the neural network has poor robustness to the lack of high frequencies).

\begin{table}[t]
	\small

	\centering
	\resizebox{0.85\linewidth}{!}{
		\begin{tabular}{ccc|ccc}
			\toprule
			\multirow{2}{*}{$r_{L}$} & \multicolumn{2}{c|}{Top-1 Acc.(\%)} & \multirow{2}{*}{$r_{L}$} & \multicolumn{2}{c}{Top-1 Acc.(\%)}                           \\
			                         & Pre-trained                         & Retrained                &                                    & Pre-trained & Retrained \\
			\hline
			145                      & 73.2                                & 73.1                     & 85                                 & 69.4        & 72.5      \\
			125                      & 73.2                                & 73.2                     & 65                                 & 63.1        & 72.1      \\
			105                      & 72.8                                & 72.9                     & 45                                 & 53.1        & 70.6      \\
			\bottomrule
		\end{tabular}
	}
	\setlength{\abovecaptionskip}{3pt}
	\caption{Comparison of full-frequency pre-trained and retrained models at different $r_{L}$.
		The model \textit{retrained from scratch} is higher than the pre-trained model at the same $r_{L}$,
		which indicates that the pre-trained model does not make full use of low- and medium-frequency information.
	}

	\label{tab:lpf_hpf}
\end{table}

\begin{figure*}[t]
	\centering
	\small
	\includegraphics[width=\linewidth]{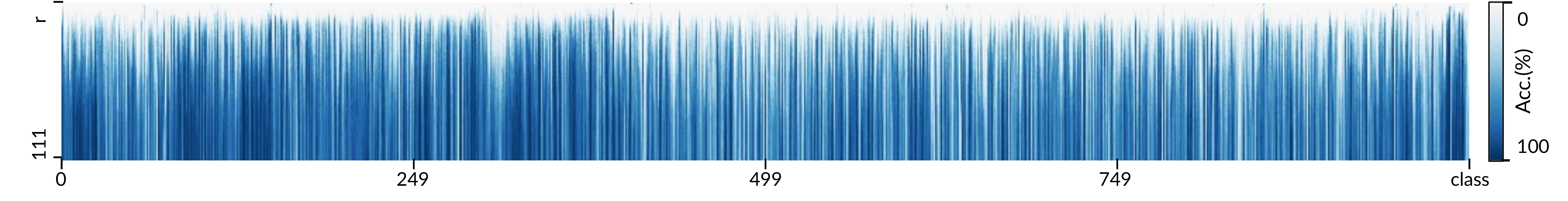}
	\setlength{\abovecaptionskip}{-15pt}
	\caption{Sensitivity of all classes.
		Most of them quickly achieve high accuracy in the low- and middle-frequency bands,
		but there are significant variations among the classes.
	}
	\label{fig:targets_freq}
\end{figure*}

\subsection{Class Level}

\begin{figure}[ht]
	\centering

	\includegraphics[width=\linewidth]{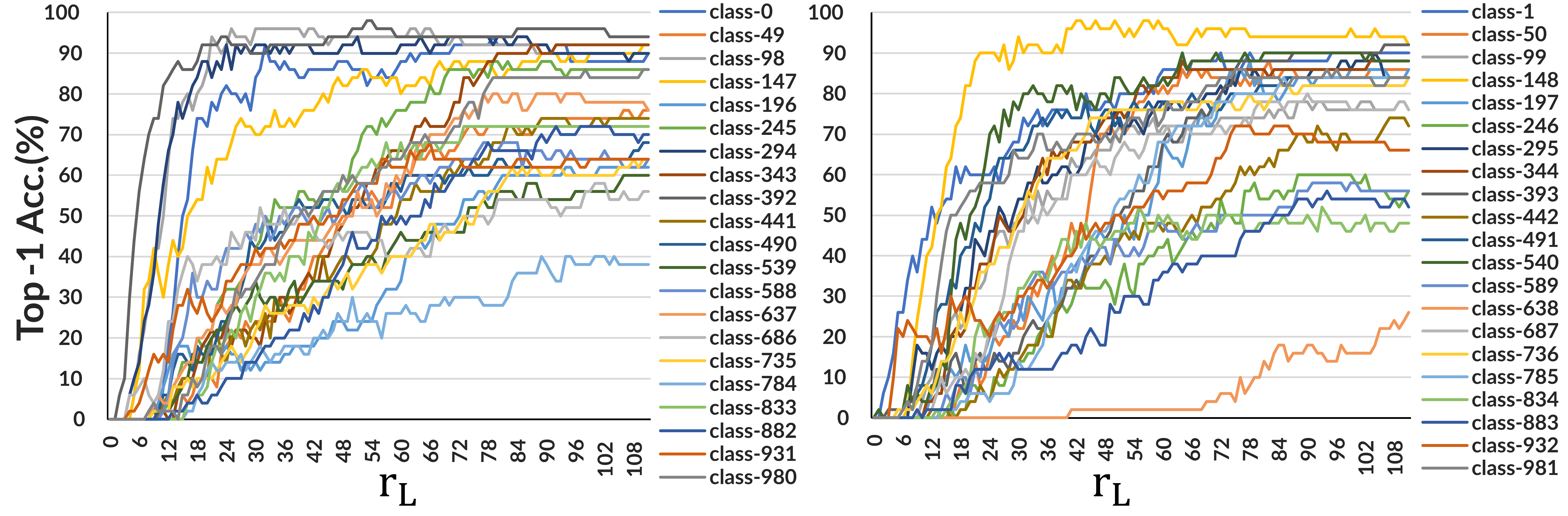}

	\setlength{\abovecaptionskip}{0pt}
	\caption{Frequency sensitivity of partial classes.
		Classes with narrow and low sensitivity bands tend to achieve better accuracy than other classes.
	}
	\label{fig:targets_freq_acc}
\end{figure}

Since the overall analysis can not capture the differences between classes, we tested the frequency sensitivity for each class.
As shown in Fig. \ref{fig:targets_freq}, most classes achieve high accuracy rapidly in low- and medium-frequency bands,
but different classes prefer different frequency bands and all frequencies are being used.
As shown in Fig. \ref{fig:targets_freq_acc}, classes with narrow and low sensitivity bands tend to achieve better accuracy than classes with high and wide sensitivity bands.
These classes often have simple and distinctive features to distinguish them from other classes, such as the classes in Fig. \ref{fig:sample_freq}a and \ref{fig:sample_freq}e.

\subsection{Sample Level}

\begin{figure}[t]
	\centering
	\includegraphics[width=\linewidth]{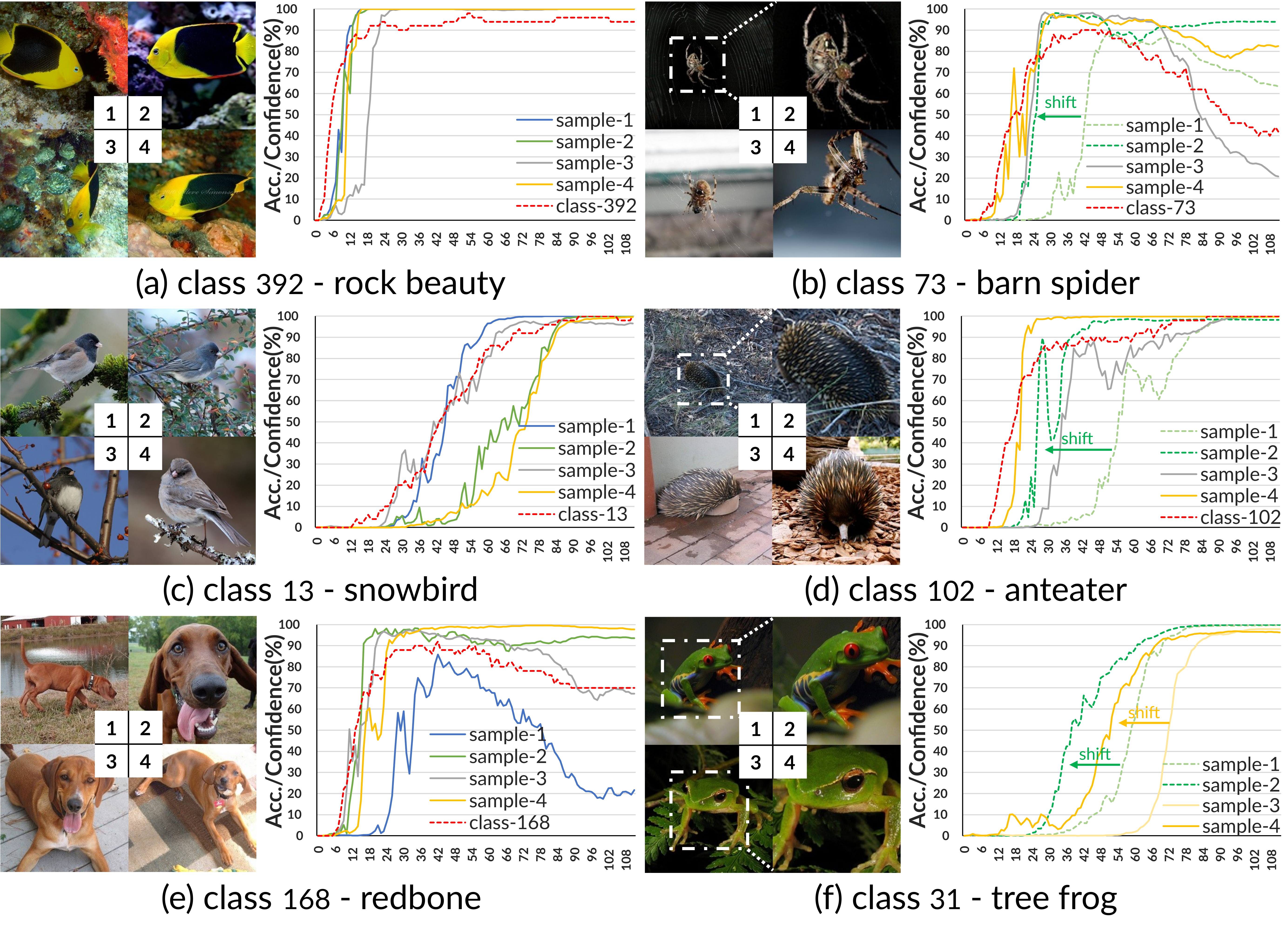}

	\setlength{\abovecaptionskip}{0pt}
	\caption{Frequency sensitivity of samples.
		The curves of the samples are similar to the curves of their classes, and the sensitive frequency bands shift with the variation of scale.
	}
	\label{fig:sample_freq}
\end{figure}

The class frequency sensitivity reflects the distribution of feature information in the frequency spectrum.
Furthermore, we tested the frequency sensitivity of the samples and the impact of scale variation on frequency sensitivity.
As shown in Fig. \ref{fig:sample_freq}, the frequency sensitivity curves of the samples and their classes are remarkably similar.
Besides there are two more points to note:
(i) the features of high-frequency components conflict with the features of the low-frequency components in some classes, as shown in Fig. \ref{fig:sample_freq}b and \ref{fig:sample_freq}e.
With the increase of high-frequency components, the accuracy of the barn spider decreases,
and its samples are gradually recognized as Phalangium opilio (another kind of spider).
(ii) the frequency sensitivity curve of the sample will shift with the variation of scale.
Sample-2 in Fig. \ref{fig:sample_freq}b is generated by enlarging and cropping sample-1 to the \textit{same} resolution,
the spider in sample-2 has a larger scale and lower frequencies.
Its curve is shifted towards lower frequencies as the scale becomes larger.
This phenomenon indicates that the frequency biases of networks are not fixed but change with the scale of the objects (Fig. \ref{fig:sample_freq}f).
In other words, the preferred frequency band depends on the scale of the features.
Consequently, neural networks must be able to extract features at various scales and frequencies.

\section{Scale-Space and GSSDNets}
\label{sec:pagestyle}

\begin{figure*}[t]
	\centering

	\includegraphics[width=0.8\linewidth]{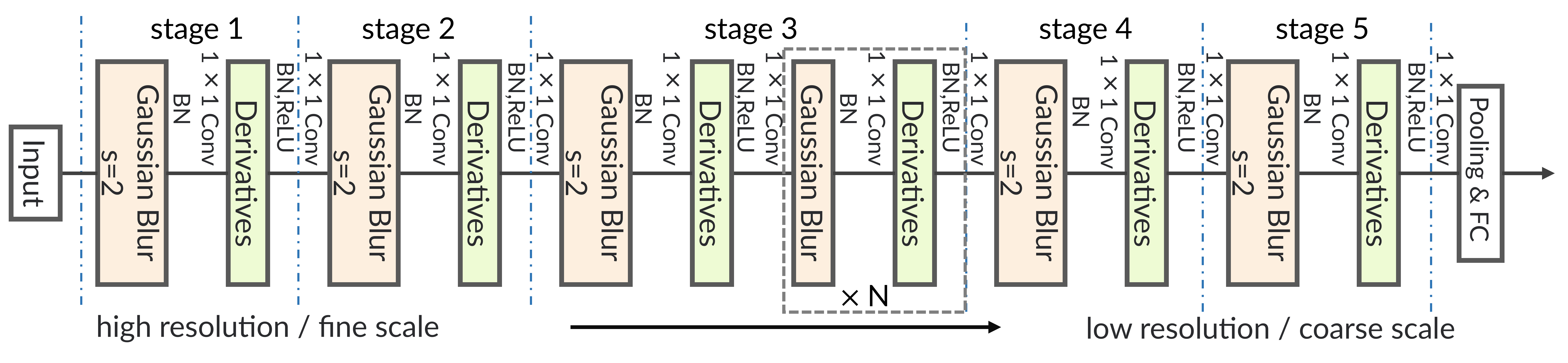}

	\setlength{\abovecaptionskip}{0pt}
	\caption{GSSDNet Architecture.
		Gaussian blurring and derivative operators are used alternately to generate multiple scales and extract features at each scale.
		These Gaussian derivatives are combined with $1 \times 1$ convolutions and nonlinearities to construct the feature extractor of GSSDNet.
	}
	\label{fig:arch}
\end{figure*}

Since neural networks do not have the scale and frequency priors,
for each input sample,
the network must attempt to extract features at all frequencies and scales needed for all classes.
To verify the importance and effectiveness of extracting features at multiple scales and frequencies,
we modify all the depthwise convolutions in DSC-based CNNs to generate a scale-space representation
and apply derivatives as local feature detectors at any scale in the scale space.

\subsection{Scale-Space and Gaussian Derivatives}

Koenderink\cite{Koenderink1984} argued that from basic principles the most simple way to observe an image at all levels of scales simultaneously is to embed the image into a one-parameter family of derived images.
And it has been shown by Koenderink\cite{Koenderink1984} and Lindeberg\cite{Lindeberg1994} that under a variety of reasonable assumptions the only possible scale-space kernel is the \textit{Gaussian function}.

For a given image $f(x, y)$, its linear \textit{scale-space representation} is a family of derived signals $L(\cdot, \cdot; s)$
defined by $L(\cdot, \cdot; 0) = f(\cdot, \cdot)$ and the convolution of $f(\cdot, \cdot)$ with the two-dimensional Gaussian kernel $g(\cdot, \cdot; s)$:
\begin{small}
	\begin{equation}
		L(\cdot, \cdot; s) = g(\cdot, \cdot; s) \ast f(\cdot, \cdot),
	\end{equation}
\end{small}where $s$ is the scale parameter.
In the linear scale-space view on local structure, taking the partial derivatives of image functions is paramount.
The \textit{scale-space derivatives} or \textit{Gaussian derivatives} at any scale $s$ are defined by
\begin{small}
	\begin{equation}
		L_{x^my^n}(\cdot, \cdot; s) = \partial_{x^my^n}L(\cdot,\cdot; s) = \partial_{x^my^n} g(\cdot, \cdot; s) \ast f(\cdot, \cdot).
	\end{equation}
\end{small}Due to the commutative property between the derivative operator and the Gaussian blurring operator,
we can either convolve the original image with Gaussian derivative operators or take the derivative at every scale in the scale space.
And we take the partial derivatives up to order 2 at every scale.

\subsection{Network Architecture}

\begin{figure}[t]
	\centering

	\subcaptionbox{Modification of the stem}[0.45\linewidth]{\includegraphics[width=0.8\linewidth]{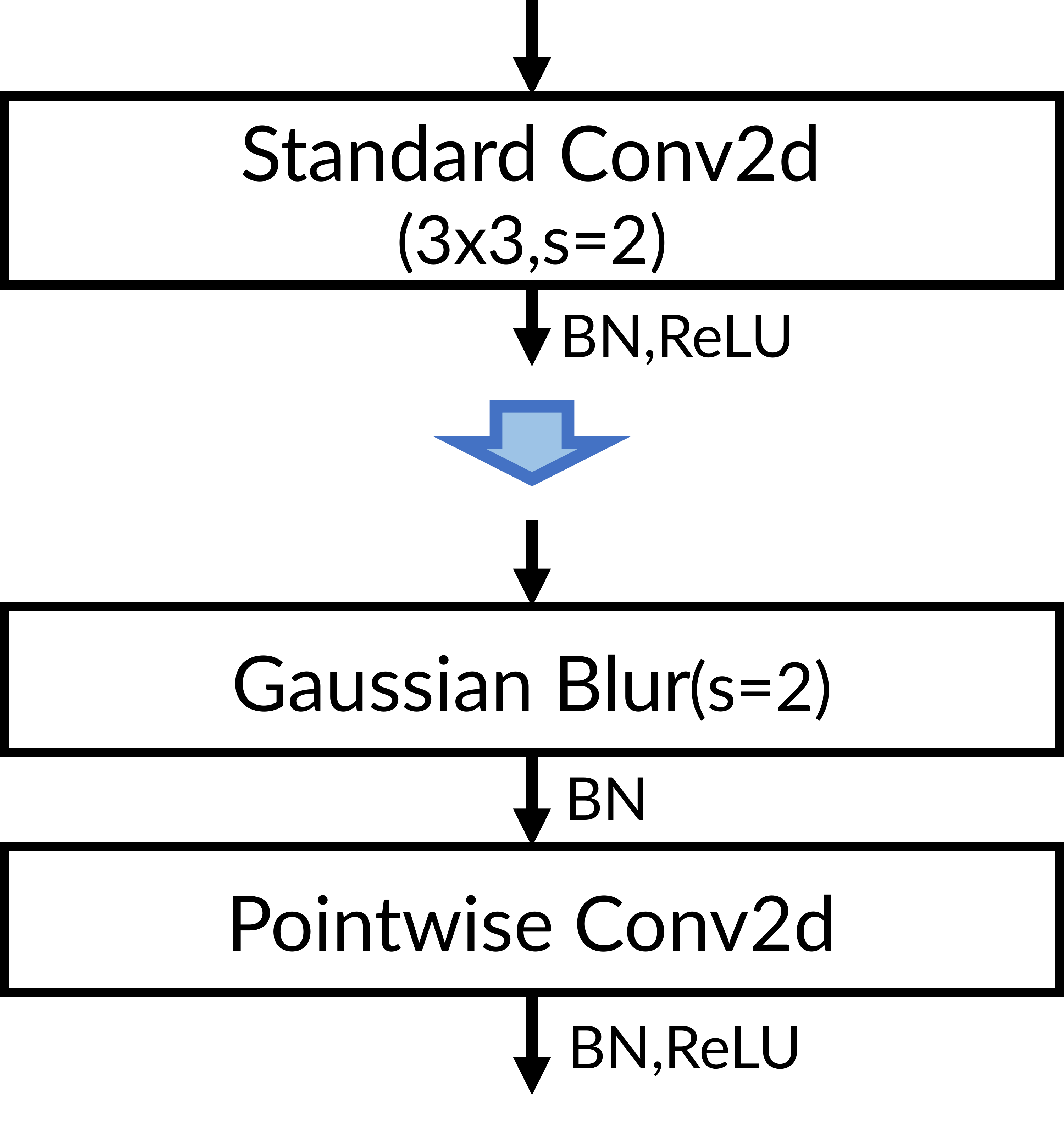}}
	\subcaptionbox{GaussianDerivatesBlock}[0.45\linewidth]{\includegraphics[width=0.8\linewidth]{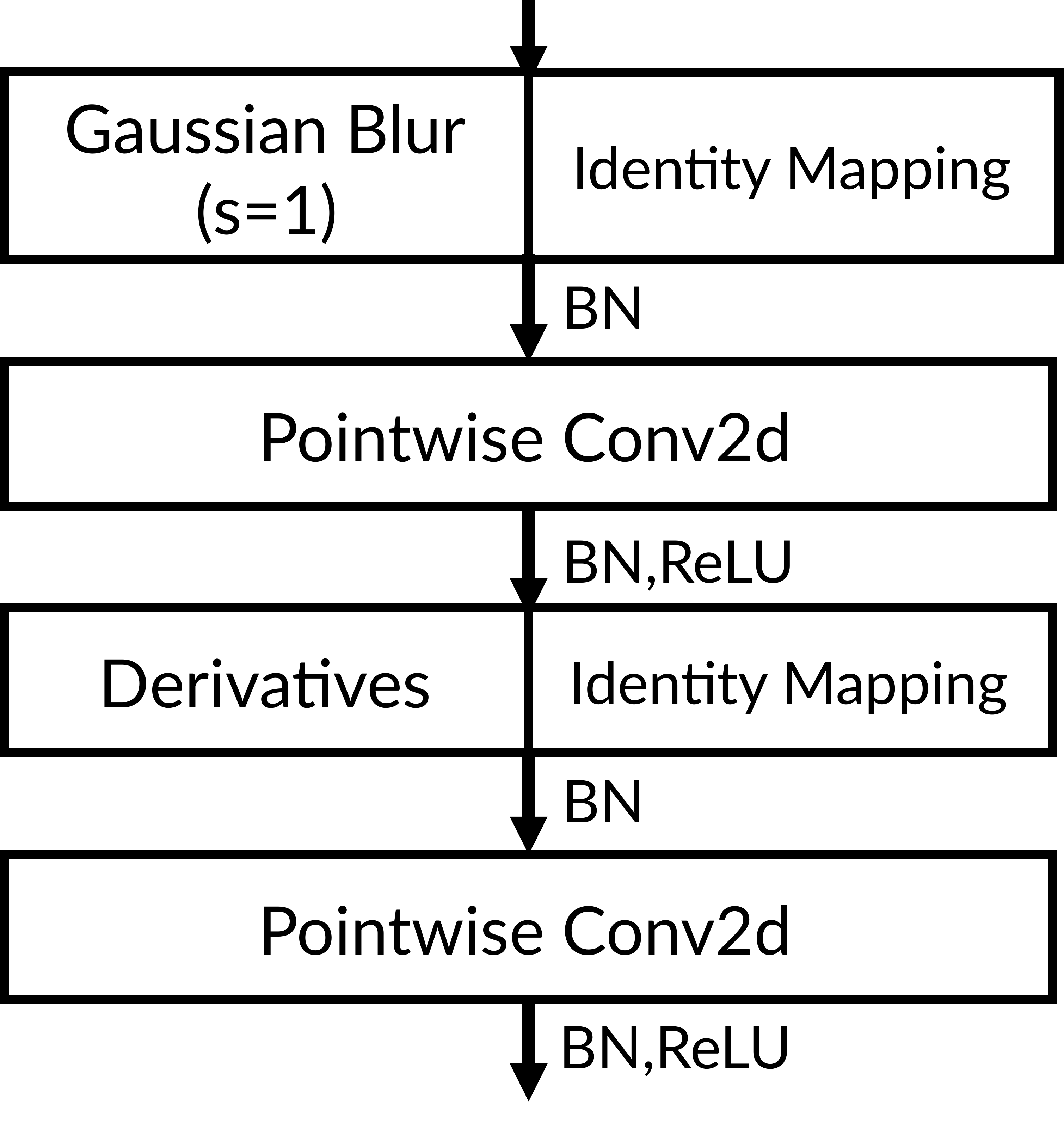}}

	\setlength{\abovecaptionskip}{3pt}
	\caption{Blocks.
		Gaussian blurring(s = 1) and derivatives only filter half of the input features. \textit{s} for stride.
	}
	\label{fig:block}
\end{figure}

The building blocks of the network are shown in Fig. \ref{fig:block}.
Compared to other vanilla DSC-based CNNs such as MobileNetV1,
we replace the first standard convolution layer with a Gaussian blurring layer and a pointwise convolution layer
and then build the \textit{GSSDNet} by stacking the \textit{GaussianDerivatesBlock}.

As shown in Fig. \ref{fig:arch},
Gaussian blurring and derivative operators are used alternately to generate coarser scales and extract local features at that scale.
These Gaussian derivatives are combined with $1 \times 1$ convolutions and nonlinearities to construct the feature extractor of GSSDNet.
Gaussian blurring and derivative operators process only half of the input features while the other half is passed directly to the next layer.
In fact, which part of the features is passed directly to the next layer is learned by pointwise convolution during the training.
Besides, the number of layers and channels of the GSSDNet is the same as that of MobileNetV1.

\section{Experiments}
\label{sec:typestyle}
In this section, we compare the GSSDNets with the \textit{VanillaNets}.
The only difference between them is that the latter are built on standard depthwise separable convolutions.

\subsection{Classification}

\noindent\textbf{Datasets}\quad ImageNet-1K\cite{Deng2010} dataset contains about 1.28M training images and 50K validation images with 1000 classes.
CIFAR-100\cite{Krizhevsky2009} consists of 50K images for training and 10K images for testing with resolution 32$\times$32.
The MNIST\cite{Yann1998} dataset contains 60K training images and 10K testing images with resolution 28$\times$28.

\noindent\textbf{Setup}\quad Our training settings are as follows:
SGD optimizer with momentum 0.9;
mini-batch size of 512;
weight decay 1e-4;
initial learning rate 0.2 with 5 warmup epochs;
batch normalization with momentum 0.9;
cosine learning rate decay for 100 epochs;
label smoothing 0.1;
the biases, $\alpha$, and $\beta$ in Batch Normalization (BN) layers are left unregularized.
Finally, all training is done on resolution $192 \times 192$.

\noindent\textbf{Results}\quad
As shown in Table \ref{tab:imagenet}, GSSDNets achieved better accuracy than the VanillaNets on MNIST and CIFAR-100 datasets,
but got a slight loss of accuracy on the ImageNet dataset.
This indicates that extracting features at different scales and frequencies by Gaussian derivatives is a very effective feature extraction method,
while vanilla CNNs are trained to learn this ability.

\begin{table}[t]
	\centering

	\resizebox{0.95\linewidth}{!}{
		\begin{tabular}{l|cc|ccc}
			\toprule
			\multirow{2}*{Model}          & \multicolumn{2}{c|}{Derivatives} & \multicolumn{3}{c}{Top-1 Acc.(\%)}                                                  \\
			                              & First                            & Second                             & MNIST          & CIFAR-100     & ImageNet      \\
			\hline
			VanillaNet $\times$1.0        &                                  &                                    & 99.68          & 76.1          & 71.8          \\
			GSSDNet $\times$1.0           & \checkmark                       &                                    & 99.69          & 74.4          & 68.5          \\
			\textbf{GSSDNet $\times$1.0}  & \checkmark                       & \checkmark                         & \textbf{99.72} & \textbf{77.1} & \textbf{71.2} \\
			\hline
			VanillaNet $\times$0.75       &                                  &                                    & 99.67          & 75.1          & 69.2          \\
			GSSDNet $\times$0.75          & \checkmark                       &                                    & 99.68          & 73.4          & 65.5          \\
			\textbf{GSSDNet $\times$0.75} & \checkmark                       & \checkmark                         & \textbf{99.69} & \textbf{76.0} & \textbf{68.4} \\
			\hline
			VanillaNet $\times$0.5        &                                  &                                    & 99.68          & 72.4          & 64.2          \\
			GSSDNet $\times$0.5           & \checkmark                       &                                    & 99.64          & 70.4          & 60.4          \\
			\textbf{GSSDNet $\times$0.5}  & \checkmark                       & \checkmark                         & \textbf{99.71} & \textbf{73.7} & \textbf{63.0} \\
			\bottomrule
		\end{tabular}
	}
	\setlength{\abovecaptionskip}{5pt}
	\caption{
		Performance results on various datasets.
		GSSDNets achieved similar accuracy as the vanilla networks.
		All GSSDNet models are scaled from the GSSDNet$\times$1.0.
	}
	\label{tab:imagenet}
\end{table}

\section{Conclusion}

This paper analyzed the CNNs from frequency and scale perspectives at overall, class, and sample levels.
At the overall level, neural networks prefer low- and medium-frequency information,
and the lack of high-frequency information only has little influence on the networks.
At the class level, different classes prefer different frequency bands,
classes with narrow and low sensitivity bands tend to achieve better accuracy than other classes.
At the sample level, the frequency biases of samples are affected by the scale of objects.
Based on these observations, we verified the hypothesis that CNNs have the fundamental ability to extract features at different scales and frequencies.
Additionally, we demonstrated that extracting features from different scales by combining Gaussian derivatives is a practical feature extraction method.

\vfill\pagebreak

\bibliographystyle{IEEEbib}
\bibliography{refs}

\end{document}